\documentclass[10pt, a4paper, english]{article}
\usepackage{lrec}
\usepackage{babel}
\usepackage{graphicx}
\usepackage{tabularx}
\usepackage{soul}
\usepackage{booktabs}
\usepackage{eurosym}
\usepackage{epstopdf}
\usepackage[latin1]{inputenc}

\usepackage{hyperref}
\usepackage{xstring}

\title{Classifying Diagrams and Their Parts using Graph Neural Networks:\\ A Comparison of Crowd-Sourced and Expert Annotations}

\name{Tuomo Hiippala}

\address{Department of Languages, University of Helsinki \\
         P.O. Box 24, 00014 University of Helsinki, Finland \\
         tuomo.hiippala@helsinki.fi \\}

\abstract{This article compares two multimodal resources that consist of diagrams which describe topics in elementary school natural sciences. Both resources contain the same diagrams and represent their structure using graphs, but differ in terms of their annotation schema and how the annotations have been created -- depending on the resource in question -- either by crowd-sourced workers or trained experts. This article reports on two experiments that evaluate how effectively crowd-sourced and expert-annotated graphs can represent the multimodal structure of diagrams for representation learning using various graph neural networks. The results show that the identity of diagram elements can be learned from their layout features, while the expert annotations provide better representations of diagram types. \\ \newline  }

\begin{document}

\maketitleabstract

\section{Introduction}

Diagrams are a common feature of many everyday media from newspapers to school textbooks, and not surprisingly, different forms of \emph{diagrammatic representation} have been studied from various perspectives. To name just a few examples, recent work has examined patterns in diagram design \cite{hullmanbach2018} and their interpretation in context \cite{tverskyetal2016}, and developed frameworks for classifying diagrams \cite{engelhardtrichards2018} and proposed guidelines for their design \cite{cheng2016}. There is also a long-standing interest in processing and generating diagrams computationally \cite{andrerist1995,batemanetal2001,batemanhenschel2007}, which is now resurfacing as advances emerging from deep learning for computer vision and natural language processing are brought to bear on diagrammatic representations \cite{sachanetal2018,choietal2018,haehnetal2019}.

From the perspective of computational processing, diagrammatic representations present a formidable challenge, as they involve tasks from both computer vision and natural language processing. On the one hand, diagrams have a \emph{spatial organisation} -- layout -- which needs to be segmented to identify meaningful units and their position. Making sense of how diagrams exploit the 2D layout space falls arguably within the domain of computer vision. On the other hand, diagrams also have a \emph{discourse structure}, which uses the layout space to set up discourse relations between instances of natural language, various types of images, arrows and lines, thus forming a unified discourse organisation. The need to parse this discourse structure shifts the focus towards the field of natural language processing.

Understanding and making inferences about the structure of diagrams and other forms of multimodal discourse may be broadly conceptualised as \emph{multimodal discourse parsing}. Recent examples of work in this area include \newcite{alikhanietal2019} and \newcite{ottoetal2019}, who model discourse relations between natural language and photographic images, drawing on linguistic theories of coherence and text--image relations, respectively. In most cases, however, predicting a single discourse relation covers only a part of the discourse structure. \newcite{sachanetal2019} note that there is a need for comprehensive theories and models of multimodal communication, as they can be used to rethink tasks that have been previously considered only from the perspective of natural language processing. 

Unlike many other areas, the study of diagrammatic representations is particularly well-resourced, as several multimodal resources have been published recently to support research on computational processing of diagrams \cite{kembhavietal2016,choietal2018,hiippalaetal2019-ai2d}. This study compares two such resources, AI2D \cite{kembhavietal2016} and AI2D-RST \cite{hiippalaetal2019-ai2d}, which both feature the same diagrams, as the latter is an extension of the former. Whereas AI2D features crowd-sourced, non-expert annotations, AI2D-RST provides multiple layers of expert annotations, which are informed by state-of-the-art approaches to multimodal communication \cite{batemanetal2017} and annotation \cite{bateman2008,hiippala2015a}. 

This provides an interesting setting for comparison and evaluation, as non-expert annotations are cheap to produce and easily outnumber the expert-annotated data, whose production consumes both time and resources. Expert annotations, however, incorporate domain knowledge from multimodality theory, which is unavailable via crowd-sourcing. Whether expert annotations provide better representations of diagrammatic structures and thus justify their higher cost is one question that this study seeks to answer.

Both AI2D and AI2D-RST represent the multimodal structure of diagrams using graphs. This enables learning their representations using graph neural networks, which are gaining currency as a graph is a natural choice for representing many types of data \cite{wuetal2019}. This article reports on two experiments that evaluate the capability of AI2D and AI2D-RST to represent the \emph{multimodal structure} of diagrams using graphs, focusing particularly on spatial layout, the hierarchical organisation of diagram elements and their connections expressed using arrows and lines.

\section{Data}

\begin{figure*}[t]
\begin{center}
\includegraphics[width=0.95\textwidth]{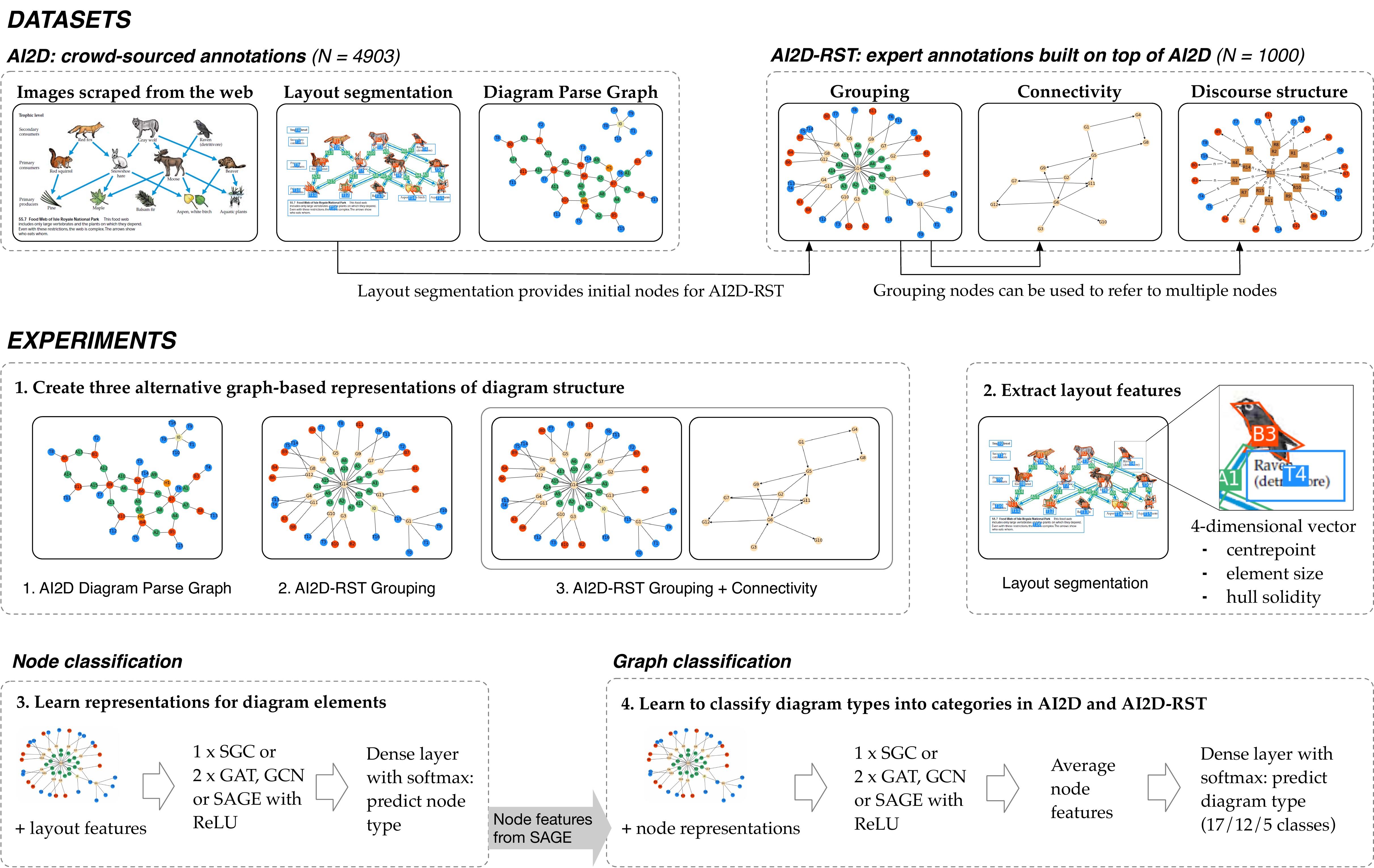} 
\caption{The relationship between crowd-sourced annotations in AI2D and AI2D-RST. AI2D-RST provides alternative, expert-annotated stand-off descriptions for a subset of 1000 diagrams from the original AI2D dataset. The grouping layer in AI2D provides a foundation for further annotation layers by allowing references to groups of nodes.}
\label{fig:datasets}
\end{center}
\end{figure*}

This section introduces the two multimodal resources compared in this study and discusses related work, beginning with the crowd-sourced annotations in AI2D and continuing with the alternative expert annotations in AI2D-RST, which are built on top of the crowd-sourced descriptions and cover a 1000-diagram subset of the original data. Figure \ref{fig:datasets} provides an overview of the two datasets, explains their relation to each other and provides an overview of the experiments reported in Section \ref{sec:experiments}

\subsection{Crowd-sourced Annotations from AI2D}

The Allen Institute for Artificial Intelligence Diagrams dataset (AI2D) contains 4903 English-language diagrams, which represent topics in primary school natural sciences, such as food webs, human physiology and life cycles, amounting to a total of 17 classes \cite{kembhavietal2016}. The dataset was originally developed to support research on diagram understanding and visual question answering \cite{kimetal2018}, but has also been used to study the contextual interpretation of diagrammatic elements, such as arrows and lines \cite{alikhanistone2018}.

The AI2D annotation schema models four types of diagram elements: text, graphics, arrows and arrowheads, whereas the semantic relations that hold between these elements are described using ten relations from a framework for analysing diagrammatic representations in \newcite{engelhardt2002}. Each diagram is represented using a Diagram Parse Graph (DPG), whose nodes stand for diagram elements while the edges between the nodes carry information about their semantic relations. The annotation for AI2D, which includes layout segmentations for the diagram images, DPGs and a multiple choice question-answer set, was created by crowd-sourced non-expert annotators on Amazon Mechanical Turk\footnote{https://www.mturk.com} \cite[243]{kembhavietal2016}.

I have previously argued that describing different types of multimodal structures in diagrammatic representations requires different types of graphs \cite{hiippalaorekhova2018}. To exemplify, many forms of multimodal discourse are assumed to possess a hierarchical structure, whose representation requires a tree graph. Diagrams, however, use arrows and lines to draw connections between elements that are not necessarily part of the same subtree, and for this reason representing connectivity requires a cyclic graph. AI2D DPGs, in turn, conflate the description of semantic relations and connections expressed using diagrammatic elements. Whether computational modelling of diagrammatic structures, or more generally, multimodal discourse parsing, benefits from pulling apart different types of multimodal structure remains an open question, which we pursued by developing an alternative annotation schema for AI2D, named AI2D-RST, which is introduced below.

\subsection{Expert Annotations from AI2D-RST}
\begin{figure}[h!]
\begin{center}
\includegraphics[scale=0.45]{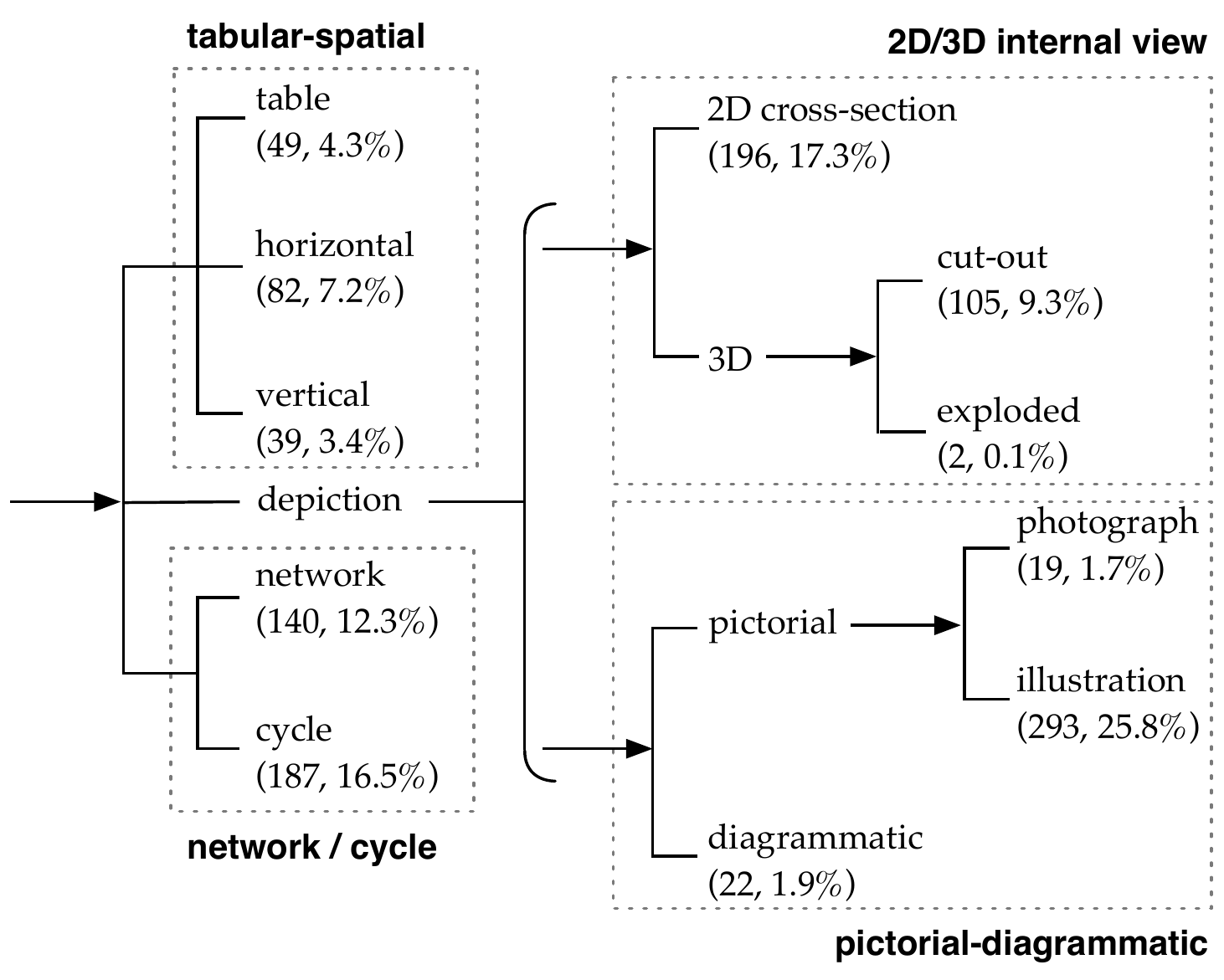} 
\caption{Fine-grained classes, their number and frequencies in AI2D-RST ($N = 1134$). Note that the number of classes exceeds the number of diagrams in AI2D-RST, as some diagrams feature multiple diagram types. The arrows indicate choices: if the diagram designer chooses depiction, a further choice must be made between pictorial--diagrammatic and 2D/3D representations. The dashed lines indicate coarse groups of diagram types.}
\label{fig:macro-groups}
\end{center}
\end{figure}

AI2D-RST covers a subset of 1000 diagrams from AI2D, which have been annotated by trained experts using a new multi-layer annotation schema for describing the diagrams in AI2D \cite{hiippalaetal2019-ai2d}. The annotation schema, which draws on state-of-the-art theories of multimodal communication \cite{batemanetal2017}, adopts a stand-off approach to describing the diagrams. Hence the three annotation layers in AI2D-RST are represented using three different graphs, which use the same identifiers for nodes across all three graphs to allow combining the descriptions in different graphs. AI2D-RST contains three graphs:
\begin{enumerate}
\item \textbf{Grouping}: A tree graph that groups together diagram elements that are likely to be visually perceived as belonging together, based loosely on Gestalt principles of visual perception \cite{ware2012}. These groups are organised into a hierarchy, which represents the organisation of content in the 2D layout space \cite{bateman2008,hiippala2015a}.

\item \textbf{Connectivity}: A cyclic graph representing connections between diagram elements or their groups, which are signalled using arrows or lines \cite{tverskyetal2000}.
\item \textbf{Discourse structure}: A tree graph representing discourse structure of the diagram using Rhetorical Structure Theory \cite{mannthompson1988,taboadamann2006a}: hence the name AI2D-RST.
\end{enumerate}
The grouping graph, which is initially populated by diagram elements from the AI2D layout segmentation, provides a foundation for describing connectivity and discourse structure by adding nodes to the grouping graph that stand for groups of diagram elements, as shown in the upper part of Figure \ref{fig:datasets}. In addition, the grouping graph includes annotations for 11 different diagram types identified in the data (e.g. cycles, cross-sections and networks), which may be used as target labels during training, as explained in Section \ref{sec:graph_classification} The coarse and fine-grained diagram types identified in the data are shown in Figure \ref{fig:macro-groups}.

\newcite{hiippalaetal2019-ai2d} show that the proposed annotation schema can be reliably applied to the data by measuring inter-annotator agreement between five annotators on random samples from the AI2D-RST corpus using Fleiss' $\kappa$ \cite{fleiss1971}. The results show high agreement on grouping ($N = 256, \kappa = 0.84$), diagram types ($N = 119, \kappa = 0.78$), connectivity ($N = 239, \kappa = 0.88$) and discourse structure ($N = 227, \kappa = 0.73$). It should be noted, however, that these measures may be affected by implicit knowledge  that tends to develop among expert annotators who work towards the same task \cite{riezler2014}.

\section{Graph-based Representations}

Both AI2D and AI2D-RST use graphs to represent the multimodal structure of diagrams. This section explicates how the graphs and their node and edge types differ across the two multimodal resources.

\subsection{Nodes}
\label{sec:nodes}
\subsubsection{Node Types}
\label{sec:node_types}
AI2D and AI2D-RST share most node types that represent different diagram elements, namely text, graphics, arrows and the image constant, which is a node that stands for the entire diagram. In AI2D, generic diagram elements such as titles describing the entire diagram are typically connected to the image constant. In AI2D-RST, the image constant acts as the root node of the tree in the grouping graph. In addition to text, graphics, arrows and the image constant, AI2D-RST features two additional node types for groups and discourse relations, whereas AI2D includes an additional node for arrowheads. To summarise, AI2D contains five distinct node types, whereas AI2D-RST has six. Note, however, that only grouping and connectivity graphs used in this study, which limits the number to five for AI2D-RST.

\subsubsection{Node Features}
\label{sec:node_features}
The same features are used for both AI2D and AI2D-RST for nodes with layout information, namely text, graphics, arrows and arrowheads (in AI2D only). The position, size and shape of each diagram element are described using the following features: (1) the centre point of the bounding box or polygon, divided by the height and width of the diagram image, (2) area, or the number of pixels within the polygon, divided by the total number of pixels in the image, and (3) the solidity of the polygon, or the polygon area divided by the area of its convex hull. This yields a 4-dimensional feature vector describing the position and size of each diagram element in the layout. Each dimension is set to zero for grouping nodes in AI2D-RST and image constant nodes in AI2D and AI2D-RST.

\subsubsection{Discourse Relations}
AI2D-RST models discourse relations using nodes, which have a 25-dimensional, one-hot encoded feature vector to represent the type of discourse relation, which are drawn from Rhetorical Structure Theory \cite{mannthompson1988}. In AI2D, the discourse relations derived from \newcite{engelhardt2002} are represented using a 10-dimensional one-hot encoded vector, which is associated with edges connecting diagram elements participating in the relation. Because the two resources draw on different theories and represent discourse relations differently, I use the grouping and connectivity graph for AI2D-RST representations and ignore the edge features in AI2D, as these descriptions attempt to describe roughly the same multimodal structures. A comparison of discourse relations is left for a follow-up study focusing on representing the discourse structure of diagrams.

\subsection{Edges}
Whereas AI2D encodes information about semantic relations using edges, in AI2D-RST the information carried by edges depends on the graph in question. The edges of the grouping graph do not have features, whereas the edges of the connectivity graph have a 3-dimensional, one-hot encoded vector that represents the type of connection. The edges of the discourse structure graph have a 2-dimensional, one-hot encoded feature vector to represent nuclearity, that is, whether the nodes that participate in a discourse relations act as nuclei or satellites.

For the experiments reported in Section 4, self-loops are added to each node in the graph. A self-loop is an edge that originates in and terminates at the same node. Self-loops essentially add the graph's identity matrix to the adjacency matrix, which allow the graph neural networks to account for the node's own features during message passing, that is, when sending and receiving features from adjacent nodes.

\section{Experiments}
\label{sec:experiments}

This section presents two experiments that compare AI2D and AI2D-RST annotations in classifying diagrams and their parts using various graph neural networks. 

\subsection{Graph Neural Networks}

I evaluated the following graph neural network architectures for both graph and node classification tasks:

\begin{itemize}
\item Graph Convolutional Network (GCN) \cite{kipfwelling2017}
\item Simplifying Graph Convolution (SGC) \cite{wuetal2019-sgc}, averaging incoming node features from up to 2 hops away
\item Graph Attention Network (GAT) \cite{velicovicetal2018} with 2 heads
\item GraphSAGE (SAGE) \cite{hamiltonetal2017} with LSTM aggregation
\end{itemize}

I implemented all graph neural networks using Deep Graph Library 0.4 \cite{wangetal2019} on the PyTorch 1.3 backend \cite{paszkeetal2017}. For GCN, GAT and SAGE, each network consists of two of the aforementioned layers with a Rectified Linear Unit (ReLU) activation, followed by a dense layer and a final softmax function for predicting class membership probabilities. For SGC, the network consists of a single SGC layer without an activation function. The implementations for each network are available in the repository associated with this article.

\subsection{Hyperparameters and Training}

I used the Tree of Parzen Estimators (TPE) algorithm \cite{bergstraetal2011} to tune model hyperparameters separately for each dataset, architecture and task using the implementation in the Tune \cite{liawetal2018} and \emph{hyperopt} \cite{bergstraetal2013} libraries. For each dataset, architecture and task, I evaluated a total of 100 hyperparameter combinations for a maximum of 100 epochs, using 850 diagrams for training and 150 for validation. The objective metric to be maximised was macro F1 score. Tables \ref{tab:hyperparams_gc} and \ref{tab:hyperparams_nc} give the hyperparameters and spaces searched for node and graph classification. Following \newcite{shcuretal2018}, I shuffled the training and validation splits for each run to prevent overfitting and used the same training procedure throughout. I used the Adam optimiser \cite{kingmaba2015} for both hyperparameter search and training. 

To address the issue of class imbalance present in both tasks, class weights were calculated by dividing the total number of samples by the product of the number of unique classes and the number of samples for each class, as implemented in \emph{scikit-learn} \cite{pedregosaetal2011}. These weights were passed to the loss function during hyperparameter search and training.

\begin{table}[h]
 \begin{center}
\begin{tabularx}{\columnwidth}{|l|X|}
\hline
Hyperparameter & Range \\
\hline
Learning rate & 0.01--0.0001 \\
\hline
Batch size & 4--32 \\
\hline
Hidden layer size & 5--30 \\
\hline
L$_2$ penalty & 0.001--0.00001 \\
\hline
\end{tabularx}
\caption{Hyperparameter ranges for graph classification} \label{tab:hyperparams_gc}
\end{center}
\end{table}

\begin{table}[h!]
 \begin{center}
\begin{tabularx}{\columnwidth}{|l|X|}
\hline
Hyperparameter & Range \\
\hline
Learning rate & 0.01--0.0001 \\
\hline
Batch size & 2--16 \\
\hline
Hidden layer size & 5--30 \\
\hline
L$_2$ penalty & 0.001--0.00001 \\
\hline
\end{tabularx}
\caption{Hyperparameter ranges for node classification} \label{tab:hyperparams_nc}
\end{center}
\end{table}

\begin{table*}[t]
\begin{tabular}{| l | r | r | r | r | r | r | r | r | r | r | r | r |}
\hline
Model & \multicolumn{3}{c|}{GAT} & \multicolumn{3}{c|}{GCN} & \multicolumn{3}{c|}{SAGE} & \multicolumn{3}{c|}{SGC} \\
\hline
Graphs &  AI2D &     G &    G+C &   AI2D &     G &   G+C &   AI2D &     G &   G+C &  AI2D &      G &    G+C \\
\hline
Accuracy    &  0.77 &  0.79 &  0.75 &  *0.81 &  0.79 &  0.78 &  *\textbf{0.93} &  0.85 &  0.88 &  0.32 &  *0.43 &  0.33 \\
Macro F1    &  0.78 &  0.77 &  0.73 &  *0.81 &  0.78 &  0.75 &  *\textbf{0.93} &  0.85 &  0.86 &  0.29 &  *0.45 &  0.28 \\
Weighted F1 &  0.77 &  0.79 &  0.76 &  *0.81 &  0.79 &  0.78 &  *\textbf{0.92} &  0.85 &  0.88 &   0.3 &  *0.42 &   0.3 \\
\hline
\end{tabular}
\caption{Mean accuracy, macro F1 and weighted F1 scores for node classification. The results are averaged over 20 runs. The following abbreviations indicate the graph used: `AI2D' for the original crowd-sourced graphs from AI2D, `G' for the grouping graph and `G+C' for the combination of grouping and connectivity graph from AI2D-RST. An asterisk indicates that the difference between AI2D and the best AI2D-RST graph is statistically significant at $p < 0.05$ when comparing the results for the given metric over 20 runs using Mann--Whitney $U$ test. The best result for each metric is marked using bold.} \label{tab:nc_results} 
\end{table*}

After hyperparameter optimisation, I trained each model with the best hyperparameter combination for 20 runs, using 850 diagrams for training, 75 for validation and 75 for testing, shuffling the splits for each run while monitoring performance on the evaluation set and stopping training early if the macro F1 score failed to improve over 15 epochs for graph classification or over 25 epochs for node classification. I then evaluated the model on the testing set and recorded the result.

\subsection{Tasks}

\subsubsection{Node Classification}
\label{sec:node_classification}

\begin{table}
\begin{tabular}{| l | r | r | r | r | r | r |}
\hline
Graphs & \multicolumn{3}{c|}{AI2D} & \multicolumn{3}{c|}{AI2D-RST (G)} \\
\hline
Model &        D &    RF &   SVM &        D &    RF &   SVM \\
\hline
Accuracy   &     0.26 &  0.76 &  0.54 &     0.25 &  0.69 &  0.62 \\
Mac. F1   &     0.20 &  0.79 &  0.60 &     0.20 &  0.56 &  0.48 \\
Wt. F1 &     0.26 &  0.76 &  0.56 &     0.25 &  0.66 &  0.58 \\
\hline
\end{tabular}
\caption{Baseline accuracy, macro F1 and weighted F1 scores for node classification from dummy (D), random forest (RF; 100 estimators) and support vector machine (SVM; $C=1.0$) classifiers with balanced class weights. The results are averaged over 20 runs. All models were implemented using \emph{scikit-learn} 0.21.3. Each node is represented by a 4-dimensional vector.}
\label{tab:nc_baseline}
\end{table}

The purpose of the \emph{node classification} task is to evaluate how well algorithms learn to classify the parts of a diagram using the graph-based representations in AI2D and AI2D-RST and node features representing the position, size and shape of the element, as described in Section \ref{sec:node_features} Identifying the correct node type is a key step when populating a graph with candidate nodes from object detectors, particularly if the nodes will be processed further, for instance, to extract semantic representations from CNN features or word embeddings. Furthermore, the node representations learned during this task can be used as node features for graph classification, as will be shown shortly below in Section \ref{sec:graph_classification}

Table \ref{tab:nc_baseline} presents a baseline for node classification from a dummy classifier, together with results for random forest and support vector machine classifiers trained on 850 and tested on 150 diagrams. Both AI2D and AI2D-RST include five node types, of which four are the same: the difference is that whereas AI2D includes arrowheads, AI2D-RST includes nodes for groups of diagram elements, as outlined in Section \ref{sec:nodes} The results seem to reflect the fact that image constants and grouping nodes have their features set to zero, and RF and SVM cannot leverage features incoming from their neighbouring nodes to learn node representations. This is likely to affect the result for AI2D-RST, which includes 7300 grouping nodes that are used to create a hierarchy of diagram elements.

Table \ref{tab:nc_results} shows the results for node classification using various graph neural network architectures. Because the results are not entirely comparable due to different node types present in the two resources, it is more reasonable to compare architectures. SAGE, GCN and GAT clearly outperform SGC in classifying nodes from both resources, as does the random forest classifier. AI2D nodes are classified with particularly high accuracy, which may result from having to learn representations for only one node type, that is, the image constant ($N = 1000$). AI2D-RST, in turn, must learn representations from scratch for both image constants ($N = 1000$) and grouping nodes ($N = 7300$).

Because SAGE learns useful node representations for both resources, as reflected in high performance for all metrics, I chose this architecture for extracting node features for graph classification.

\subsubsection{Graph Classification}
\label{sec:graph_classification}

This task compares the performance of graph-based representations in AI2D and AI2D-RST for \emph{classifying entire diagrams}. Here the aim is to evaluate to what extent graph neural networks can learn about the generic structure of primary school science diagrams from the graph-based representations in AI2D and AI2D-RST. Correctly identifying what the diagram attempts to communicate and \emph{how} carries implications for tasks such as visual question answering, as the type of a diagram constrains the interpretation of key diagrammatic elements, such as the meaning of lines and arrows \cite{tverskyetal2016,alikhanistone2018}.

\begin{table*}
\begin{tabular}{| l | r | r | r | r | r | r | r | r | r | r | r | r |}
\hline
Classes & \multicolumn{12}{c|}{Original classes from AI2D ($N = 17$)} \\
\hline
Model & \multicolumn{3}{c|}{GAT} & \multicolumn{3}{c|}{GCN} & \multicolumn{3}{c|}{SAGE} & \multicolumn{3}{c|}{SGC} \\
\hline
Graphs &     AI2D &     G &    G+C &  AI2D &     G &    G+C &  AI2D &     G &    G+C &  AI2D &     G &   G+C \\
\hline
Accuracy    &     0.53 &  $^+$0.54 &  0.49 &  0.52 &  $^+$0.55 &   0.5 &  *\textbf{0.58} &   0.5 &  0.52 &  0.55 &   0.54 &  0.53 \\
Macro F1    &     0.26 &   0.25 &  0.26 &  0.23 &  $^+$0.27 &  0.23 &  *\textbf{0.32} &  0.25 &  0.26 &  0.23 &  $^+$0.22 &  0.19 \\
Weighted F1 &     0.53 &   0.56 &  0.52 &   0.5 &   0.53 &  0.51 &   *\textbf{0.6} &  0.54 &  0.54 &   0.5 &  $^+$0.48 &  0.44 \\
\hline
Classes & \multicolumn{12}{c|}{Coarse classes from AI2D-RST ($N = 5$)} \\
\hline
Model & \multicolumn{3}{c|}{GAT} & \multicolumn{3}{c|}{GCN} & \multicolumn{3}{c|}{SAGE} & \multicolumn{3}{c|}{SGC} \\
\hline
Graphs &   AI2D &      G &    G+C &  AI2D &      G &    G+C &  AI2D &     G &   G+C &  AI2D &     G &   G+C \\
\hline
Accuracy    &   0.59 &   $^+$0.61 &  0.58 &   0.6 &  *\textbf{0.63} &  0.61 &   0.6 &  0.58 &    0.6 &  *0.56 &   0.5 &  0.51 \\
Macro F1    &   0.46 &  *$^+$\textbf{0.51} &  0.47 &  0.46 &   *0.5 &  0.47 &  0.47 &  0.48 &  *0.49 &   0.41 &  0.39 &   0.4 \\
Weighted F1 &   0.56 &   *$^+$0.6 &  0.55 &  0.58 &    \textbf{0.6} &  0.58 &  0.57 &  0.56 &   0.58 &    0.5 &  0.47 &  0.48 \\
\hline
Classes & \multicolumn{12}{c|}{Fine-grained classes from AI2D-RST ($N = 12$)} \\
\hline
Model & \multicolumn{3}{c|}{GAT} & \multicolumn{3}{c|}{GCN} & \multicolumn{3}{c|}{SAGE} & \multicolumn{3}{c|}{SGC} \\
\hline
Graphs &         AI2D &     G &    G+C &  AI2D &     G &    G+C &  AI2D &     G &    G+C &  AI2D &     G &   G+C \\
\hline
Accuracy    &         0.39 &  0.36 &  0.37 &   0.4 &  0.39 &   0.42 &  0.42 &  0.41 &  \textbf{0.43} &  0.39 &  0.31 &  $^+$0.36 \\
Macro F1    &         0.27 &  0.25 &  0.25 &  0.24 &  0.26 &  \textbf{*0.29} &  0.28 &  0.28 &  0.29 &  0.22 &  0.19 &  $^+$0.21 \\
Weighted F1 &         0.36 &  0.34 &  0.34 &  0.35 &  0.38 &  *0.39 &  0.39 &  0.38 &   \textbf{0.4} &  0.31 &  0.27 &  $^+$0.31 \\

\hline
\end{tabular}

\caption{Mean accuracy, macro F1 and weighted F1 scores for graph classification. The results are averaged over 20 runs. The following abbreviations indicate the graph used: `AI2D' for the original crowd-sourced graphs from AI2D, `G' for the grouping graph and `G+C' for the combination of grouping and connectivity graph from AI2D-RST. * indicates that the difference between AI2D and the best AI2D-RST graph is statistically significant at $p < 0.05$ when comparing the results over 20 runs for the given metric using Mann--Whitney $U$ test. $^+$ indicates the same for AI2D-RST grouping graph and the combination of grouping and connectivity graphs. The best result for each metric across all models and graphs is marked in bold.} \label{tab:gc_results}
\end{table*}

\begin{figure}
\begin{center}
\includegraphics[scale=0.45]{4210.png} 
\caption{Diagram \#4120 in AI2D combines two diagram types: a \emph{cross-section} with a \emph{cycle} (cf. Figure \ref{fig:macro-groups})}
\label{fig:mixed}
\end{center}
\end{figure}

To enable a fair comparison, the target classes are derived from both AI2D and AI2D-RST. Whereas AI2D includes 17 classes that represent the \emph{semantic content} of diagrams, as exemplified by categories such as `parts of the Earth', `volcano', and `food chains and webs', AI2D-RST classifies diagrams into abstract \emph{diagram types}, such as cycles, networks, cross-sections and cut-outs. More specifically, AI2D-RST provides classes for diagram types at two levels of granularity, fine-grained (12 classes) and coarse (5 classes), which are derived from the proposed schema for diagram types in AI2D-RST \cite{hiippalaetal2019-ai2d}.

The 11 fine-grained classes in AI2D-RST shown in Figure \ref{fig:macro-groups} are complemented by an additional class (`mixed'), which includes diagrams that combine multiple diagram types, whose inclusion avoids performing multi-label classification (see the example in Figure \ref{fig:mixed}). The coarse classes, which are derived by grouping fine-grained classes for tables, tabular and spatial organisations, networks and cycles, diagrammatic and pictorial representations, and so on, are also complemented by a `mixed' class.

For this task, the node features consist of the representations learned during node classification in Section \ref{sec:node_classification} These representations are extracted by feeding the features representing node position, size and shape to the graph neural network, which in both cases uses the GraphSAGE architecture \cite{hamiltonetal2017}, and recording the output of the final softmax activation. Compared to a one-hot encoding, representing node identity using a probability distribution from a softmax activation reduces the sparsity of the feature vector. This yields a 5-dimensional feature vector for each node.

Table \ref{tab:gc_baseline} provides a baseline for graph classification from a dummy classifier, as well as results for random forest (RF) and support vector machine (SVM) classifiers trained on 850 and tested on 150 diagrams. The macro F1 scores show that the RF classifier with 100 decision trees offers competitive performance for all target classes and both AI2D and AI2D-RST, in some cases outperforming graph neural networks. It should be noted, however, that the RF classifier is trained with node features learned using GraphSAGE.

\begin{table}[h]
\begin{tabular}{| l | r | r | r | r | r | r |}
\hline
Classes & \multicolumn{6}{c|}{AI2D ($N = 17$)} \\
\hline
Dataset & \multicolumn{3}{c|}{AI2D} & \multicolumn{3}{c|}{AI2D-RST (G)} \\
\hline
Model &        D &    RF &   SVM &        D &    RF &   SVM \\
\hline
Acc.   &     0.15 &  0.59 &  0.23 &     0.15 &  0.58 &  0.30 \\
Mac. F1   &     0.06 &  0.29 &  0.13 &     0.06 &  0.24 &  0.10 \\
Wt. F1 &     0.15 &  0.55 &  0.19 &     0.15 &  0.54 &  0.28 \\
\hline
Classes & \multicolumn{6}{c|}{AI2D-RST (coarse) ($N = 5$)} \\
\hline
Dataset & \multicolumn{3}{c|}{AI2D} & \multicolumn{3}{c|}{AI2D-RST (G)} \\
\hline
Model &      D &    RF &   SVM &        D &    RF &   SVM \\
\hline
Acc.   &   0.24 &  0.61 &  0.53 &     0.25 &  0.61 &  0.43 \\
Mac. F1   &   0.19 &  0.47 &  0.38 &     0.20 &  0.47 &  0.34 \\
Wt. F1 &   0.24 &  0.58 &  0.46 &     0.24 &  0.59 &  0.40 \\
\hline
Classes & \multicolumn{6}{c|}{AI2D-RST (fine-grained) ($N=12$)} \\
\hline
Dataset & \multicolumn{3}{c|}{AI2D} & \multicolumn{3}{c|}{AI2D-RST (G)} \\
\hline
Model &     D &    RF &   SVM &        D &    RF &   SVM \\
\hline
Acc.   &  0.15 &  0.45 &  0.26 &     0.13 &  0.44 &  0.26 \\
Mac. F1   &  0.09 &  0.28 &  0.14 &     0.08 &  0.25 &  0.13 \\
Wt. F1 &  0.15 &  0.42 &  0.19 &     0.14 &  0.41 &  0.20 \\
\hline
\end{tabular}
\caption{Baseline accuracy, macro F1 and weighted F1 scores for graph classification using dummy (D), random forest (RF; 100 estimators) and support vector machine (SVM; $C=1.0$) classifiers with balanced class weights. The results are averaged over 20 runs. All models were implemented using \emph{scikit-learn} 0.21.3. Each diagram is represented by a 5-dimensional vector acquired by averaging the features for all nodes in the graph.}
\label{tab:gc_baseline}
\end{table}

The results for graph classification using graph neural networks presented in Table \ref{tab:gc_results} show certain differences between AI2D and AI2D-RST. When classifying diagrams into the original semantic categories defined in AI2D ($N = 17$), the AI2D graphs significantly outperform AI2D-RST when using the GraphSAGE architecture. For all other graph neural networks, the differences between AI2D and AI2D-RST are not statistically significant. This is not surprising as the AI2D graphs were tailored for the original classes, yet the AI2D-RST graphs seem to capture generic properties that help to classify diagrams into semantic categories nearly as accurately as AI2D graphs designed specifically for this purpose, although no semantic features apart from the layout structure are provided to the classifier.

The situation is reversed for the coarse ($N = 5$) and fine-grained ($N = 12$) classes from AI2D-RST, in which the AI2D-RST graphs generally outperform AI2D, except for coarse classification using SGC. This classification task obviously benefits AI2D-RST, whose classification schema was originally designed for abstract diagram types. This may also suggest that the AI2D graphs do not capture regularities that would support learning to generalise about diagram types. The situation is somewhat different for fine-grained classification, in which the differences in performance are relatively small.

Generally, most architectures do not benefit from combining the grouping and connectivity graphs in AI2D-RST. This is an interesting finding, as many diagram types differ in terms of their connectivity structures (e.g. cycles and networks) \cite{hiippalaetal2019-ai2d}. The edges introduced from the connectivity graph naturally increase the flow of information in the graph, but this does not seem to help learn distinctive features between diagram types. On the other hand, it should be noted that the nodes are not typed, that is, the model cannot distinguish between edges from the grouping and connectivity graphs.

Overall, the macro F1 scores for both AI2D and AI2D-RST, which assigns equal weight to all classes regardless of the number of samples, underline the challenge of training classifiers using limited data with imbalanced classes. The lack of visual features may also affect overall classification performance: certain fine-grained classes, which are also prominent in the data, such as 2D cross-sections and 3D cut-outs, may have similar graph-based representations. Extracting visual features from diagram images may help to discern between diagrams whose graphs bear close resemblance to one another, but this would require advanced object detectors for non-photographic images.

% ADD t-SNE of macro-groups here!

\section{Discussion}

% When paired with a suitable graph neural network architecture such as GraphSAGE \cite{hamiltonetal2017}, the expert-annotated AI2D-RST graphs outperform their crowd-sourced AI2D counterparts in most tasks. Nevertheless, whether an average improvement of X.X points for the macro F1 scores justifies the cost of approximately 50000\euro{} used to develop AI2D-RST is questionable. The impact of a resource, however, can also be evaluated in more than just terms of improved performance. In this case, attention should be paid to the improvements gained by introducing annotations informed by theories of multimodal communication and diagrammatic representation.

The results for AI2D-RST show that the grouping graph, which represents visual perceptual groups of diagram elements and their hierarchical organisation, provides a robust foundation for describing the spatial organisation of diagrammatic representations. This kind of generic schema can be expanded \emph{beyond} diagrams to other modes of expression that make use of the spatial extent, such as entire page layouts. A description of how the layout space is used can be incorporated into any effort to model discourse relations that may hold between the groups or their parts.

The promising results AI2D-RST suggest is that domain experts in multimodal communication should be involved in planning crowd-sourced annotation tasks right from the beginning. Segmentation, in particular, warrants attention as this phase defines the units of analysis: cut-outs and cross-sections, for instance, use labels and lines to pick out sub-regions of graphical objects, whereas in illustrations the labels often refer to entire objects. Such distinctions should preferably be picked out at the very beginning to be incorporated fully into the annotation schema.

Tasks related to grouping and connectivity annotation could be crowd-sourced relatively easily, whereas annotating diagram types and discourse relations may require multi-step procedures and assistance in the form of prompts, as \newcite{yungetal2019} have recently shown for RST. Involving both expert and crowd-sourced annotators could also alleviate problems related to circularity by forcing domain experts to frame the tasks in terms understandable to crowd-sourced workers \cite{riezler2014}.

In light of the results for graph classification, one should note that node features are averaged before classification \emph{regardless of their connections in the graph}. Whereas the expert-annotated grouping graph in AI2D-RST has been pruned from isolated nodes, which ensures that features are propagated to neighbouring nodes, the crowd-sourced AI2D graphs contain both isolated nodes and subgraphs. To what extent these disconnections affect the performance for AI2D warrant a separate study. Additionally, more advanced techniques than mere averaging, such as pooling, should be explored in future work.

Finally, there are many aspects of diagrammatic representation that were not explored in this study. To begin with, a comparison of representations for discourse structures using the question-answering set accompanying AI2D would be particularly interesting, especially if both AI2D and AI2D-RST graphs were enriched with features from state of the art semantic representations for natural language and graphic elements.

\section{Conclusion}

In this article, I compared graph-based representations of diagrams representing primary school science topics from two datasets that contain the same diagrams, which have been annotated by either crowd-sourced workers or trained experts. The comparison involved two tasks, graph and node classification, using four different architectures for graph neural networks, which were compared to baselines from dummy, random forest and support vector machine classifiers.

The results showed that graph neural networks can learn to accurately identify diagram elements from their size, shape and position in layout. These node representations could then be used as features for graph classification. Identifying diagrams, either in terms of what they represent (semantic content) or how (abstract diagram type), proved more challenging using the graph-based representations. Improving accuracy may require additional features that capture visual properties of the diagrams, as these distinctions cannot be captured by graph-based representations and features focusing on layout.

Overall, the results nevertheless suggest that simple layout features can provide a foundation for representing diagrammatic structures, which use the layout space to organise the content and set up discourse relations between different elements. To what extent these layout features can support the prediction of actual discourse relations should be explored in future research.

\section{Bibliographical References}
\label{main:ref}

\bibliographystyle{lrec}
\bibliography{bibliography}

\end{document}